# National level satellite-based crop field inventories in smallholder landscapes

Philippe Rufin*[1,2], Pauline Lucie Hammer[2], Leon-Friedrich Thomas[2,3], Sá Nogueira Lisboa[4,5], Natasha Ribeiro[4], Almeida Sitoe[4], Patrick Hostert[2], Patrick Meyfroidt[1,6]

[1] Earth and Life Institute, UCLouvain, Place Pasteur 3, 1348 Louvain-la-Neuve, Belgium

[2] Geography Department, Humboldt-Universität zu Berlin, Unter den Linden 6, 10117 Berlin, Germany

[3] Department of Agricultural Sciences, University of Helsinki, P.O. Box 28, FI-00014 Helsinki, Finland

[4] Faculty of Agronomy and Forest Engineering, Eduardo Mondlane University, PO Box 257, Maputo, Mozambique

[5] N'Lab, Nitidæ, Maputo, Mozambique.

[6] F.R.S.-FNRS, Rue d'Egmont 5, 1000 Brussels, Belgium

* corresponding author: philippe.rufin@uclouvain.be

## Significance

Smallholder farms are strongholds of global food production and provide livelihoods to rural populations across the globe. Surprisingly, fundamental properties of smallholder systems remain only weakly understood, which hampers the design of sustainability-centered policies and aid interventions. We here present an approach to delineate individual agricultural fields from very-high resolution satellite imagery at the national level. We derive detailed field boundaries for 21 million fields in Mozambique, allowing unique insights into the spatial distribution of agriculture, field size and linkages between agriculture and forest cover change. The workflow has minimum reference data requirements and is highly transferable across regions, providing a stepping-stone for better understanding smallholder farming systems.

# Abstract

The design of science-based policies to improve the sustainability of smallholder agriculture is challenged by a limited understanding of fundamental system properties, such as the spatial distribution of active cropland and field size. We integrate very high spatial resolution (1.5 m) Earth observation data and deep transfer learning to derive crop field delineations in complex agricultural systems at the national scale, while maintaining minimum reference data requirements and enhancing transferability. We provide the first national-level dataset of 21 million individual fields for Mozambique (covering ~800,000 km²) for 2023. Our maps separate active cropland from non-agricultural land use with an overall accuracy of 93% and balanced omission and commission errors. Field-level spatial agreement reached median intersection over union (IoU) scores of 0.81, advancing the state-of-the-art in large-area field delineation in complex smallholder systems. The active cropland maps capture fragmented rural regions with low cropland shares not yet identified in global land cover or cropland maps. These regions are mostly located in agricultural frontier regions which host 7-9% of the Mozambican population. Field size in Mozambique is very low overall, with half of the fields being smaller than 0.16 ha, and 83% smaller than 0.5 ha. Mean field size at aggregate spatial resolution (0.05°) is 0.32 ha, but it varies strongly across gradients of accessibility, population density, and net forest cover change. This variation reflects a diverse set of actors, ranging from semi-subsistence smallholder farms to medium-scale commercial farming, and large-scale farming operations. Our results highlight that field size is a key indicator relating to socio-economic and environmental outcomes of agriculture (e.g., food production, livelihoods, deforestation, biodiversity), as well as their trade-offs.

# Keywords

# Introduction

Smallholder farms - commonly defined as farms operating on 2 hectares of land or less - are cornerstones of global food production and biodiversity (Ricciardi et al., 2018, 2021; Samberg et al., 2016). At the same time, smallholder-dominated regions, especially in the tropics, are particularly susceptible to undernutrition and poverty, currently amplified by declining productivity trends which pose a threat to rural populations, particularly across the African continent (Wollburg et al., 2024). This trend is alarming, as smallholder farming constitutes a key livelihood in regions with rapidly growing populations and high exposure to extreme events in a changing climate (Morton, 2007).

Timely and spatially detailed knowledge on farming system properties is essential to inform efforts targeting the reduction of deforestation and forest degradation, development and aid interventions for poverty alleviation, and for monitoring the progress of international commitments such as the United Nation´s Sustainable Development Goals (SDG). However, research and interventions in smallholder farming systems are hampered by empirical blind spots resulting from a lack of harmonized subnational data collection efforts. Despite recent progress (Lee et al., 2025), basic properties such as the distribution of actively used cropland, field size (i.e., the area of individual fields), or farm size (i.e., the total area of all fields operated by a farm) remain unknown. The geographic coverage and timeliness of key SDG indicators remain limited, and resource constraints hamper data collection, particularly in lower- and lower-middle income countries (LLMICs) (United Nations Department of Economic and Social Affairs, 2023).

Satellite-based Earth observation (EO) has been identified as a high priority tool to support national SDG reporting (United Nations Department of Economic and Social Affairs, 2023). EO has the potential to inform 34 targets across 11 of the 17 SDGs, including indicators for target 2.3 (double the agricultural productivity and incomes of small-scale food producers) and 2.4 (ensure sustainable food production systems) (O'Connor et al., 2020) but reliable EO-based indicators are typically not available for most LLMICs and smallholder dominated regions.

Publicly available EO data or derived global scale products are often not fit for purpose in smallholder systems. This issue arises from the spatial scale of farming operations, high degrees of landscape fragmentation, locally variable practices of land management, and rapid change dynamics in smallholder farming systems (Rufin et al., 2022; Wei et al., 2020). Consequently, data of high timeliness and very high spatial resolution is needed – depending on the application below 5 m, sometimes finer than 2 m. However, limited and costly access to commercial very-high-resolution satellite imagery continues to hamper the operational production of relevant SDG indicators in smallholder landscapes (Nakalembe & Kerner, 2023; Rufin et al., 2025).

Satellite-based crop field delineation – the task of identifying and accurately delineating individual management units from satellite imagery – is among the Essential Agricultural Variables (EAV) identified as crucial for SDG monitoring by the Group on Earth Observations Global Agricultural Monitoring Initiative (GEOGLAM) (Whitcraft et al., 2019). Field delineation in smallholder agriculture allows for precise monitoring of indicators linked to SDG 2.3 and 2.4 and facilitates field-scale analyses on the types and productivity of crops, or how land is managed. Also, at aggregate (e.g., national) scale, such data can help to target development interventions or facilitate the assessment of disaster impacts, food security, rural livelihoods, or structural transitions in agriculture, which manifest in changing field size.

Field size is expected to vary across environmental gradients (e.g., soil properties and climate), socio-economic contexts (e.g., access to markets, labor opportunities, the availability of inputs), and institutional settings (e.g., through agricultural and land tenure policies). Moreover, spatially detailed farm size estimates can be derived from field size data when accounting for socio-economic and environmental properties (Jänicke et al., 2024). The link between field size and farm size further suggests that field size relates to actors operating the farm, including various categories of smallholders as well as medium-scale commercial farmers that emerge in many regions across sub-Saharan Africa (Jayne et al., 2019). Since such shifts in actors may reflect land consolidation processes, field size dynamics can inform on locally prevalent development trajectories.

Beyond its relevance for food production, agriculture is also the major driver of tropical deforestation (Pendrill et al., 2022). The types and magnitudes of such land change processes differ by actors. In Africa, small-scale agriculture - as opposed to large-scale farming operations – has been flagged as the major driver of deforestation (Branthomme et al., 2023). However, more detailed data on field size allows to move beyond this simple small-versus-large dichotomy by identifying medium-scale farms as separate actors driving forest cover loss through their own set of causal mechanisms (Wineman et al., 2022). Engagement in the production of commodity crops, market-oriented production, and access to agricultural inputs, render medium-scale farmers a distinct set of actors that should be differentiated from (semi-)subsistence smallholder farmers and large-scale agro-industrial operations (Kwapong et al., 2021; Sitko & Chamberlin, 2015).

Spatially detailed knowledge about field size can thus enhance our understanding of smallholder-dominated farming systems, but the required data remains scarce. While one global dataset on field size exists (Lesiv et al., 2019), it is limited in time (providing a single snapshot in time around 2010) and represents field size in broad categories, with the smallest category "very small field" containing all fields <0.64 ha. Around 50% of the African territory has been assigned to this category, hiding important variations within smallholder systems where most fields are substantially smaller. Beyond this global dataset, national-scale field delineation datasets do not commonly exist in smallholder landscapes (an exception being Ghana (Estes et al., 2022)), and particularly not for multiple points in time, as suggested by GEOGLAM in the context of SDG monitoring (Whitcraft et al., 2019).

We here provide the first national-scale field delineation dataset derived from 1.5 m spatial resolution satellite imagery. We use a state-of-the-art deep learning architecture paired with a series of machine-learning post-processing methods for crop field delineation. Our methods build on openly available pre-trained models that can be transferred across geographies with minimum reference data requirements (Rufin et al., 2024), making the approach highly suitable for smallholder farming contexts where reference data commonly remains scarce (Nakalembe & Kerner, 2023). We selected Mozambique as a case study to develop the approach due to its high reliance on agriculture, which is the main economic activity for 85% of rural households (World Bank, 2022). Most farms use only manual labor and little to no agricultural inputs, which results in a high landscape complexity, rendering Mozambique a highly challenging context for satellite-based field delineation.

We present *MozFields 2023* – a national-level field delineation dataset containing approximately 21 million individual fields for the target year 2023. We define field as a unit of land which can be delineated by its physical appearance through markers of land management or land tenure and which is actively used to produce annual or (semi-)perennial crops at the time of observation. Our definition excludes fallows, tree crops and agroforestry systems. We compare the cropland distribution in *MozFields 2023* to existing global scale maps and highlight how our map provides a unique perspective of the agricultural landscape with respect to the extent of actively used cropland. Our continuous maps of field size indicators reveal nuanced spatial patterns and show how these reflect the proportions of actors present in the Mozambican farming landscape. Moreover, we reveal how field size varies across socio-economic contexts (population density and accessibility). Lastly, we assess field size against net forest cover change in the years 2010 to 2020, providing novel insights into the drivers of forest cover loss.

# Results

### *National-scale mapping of individual fields at unprecedented detail*

Our approach uses a state-of-the-art deep learning model for field delineation based on 1.5 m resolution imagery. This allows us to produce detailed, accurate, wall-to-wall maps of cropland and field delineations over large extents. We leverage transfer learning with pseudo-labels (Rufin et al., 2024) to fine-tune model weights from a pre-trained field delineation model (Wang et al., 2022) across geographies (here from India to Mozambique). The approach limits the need for manually annotated reference data to the data employed during validation (see Fig. S1).

The map of individual fields covers the entire territory of Mozambique (782,477 km², excluding Lake Malawi), for the target year 2023. We mapped actively cropped fields with an overall pixel-level accuracy of 93% and balanced user´s and producer´s accuracies in the non-cropland (96%, 90%) and active cropland domain (61%, 64%). Errors in the cropland domain remain moderately high compared to generic LULC maps. This is in parts driven by the very high spatial detail of our image data, which may affect the distinction between cropland and non-cropland land cover. Additionally, identifying active cropland involves confusion with short-term fallows (Tong et al., 2020), contributing to the error. We found that commission and omission errors dominantly cluster in regions with relatively high shares of cropland, where fallow fractions tend to be relatively high (Rufin et al., 2022). Importantly, our approach does not rely on ancillary cropland masks as common in field delineation studies in other world regions, where such data is available. Yet, for regions where accurate and timely maps of active cropland are available, these may constitute a valuable input for further improving the thematic accuracy of the field delineation.

The spatial agreement of individual fields was found to be high with mean and median intersection over union (IoU) scores of 0.76 and 0.81. More than 92% of the reference fields were identified with an IoU above 0.5, and 53% of the fields with an IoU above 0.8. While previous work has obtained slightly higher IoU scores using SPOT data in India (median IoU 0.89 in Wang et al. (2022) vs. median IoU 0.81 in *MozFields* 2023), the difference likely arises from to the more consolidated agricultural systems in India: The Indian farming systems features larger, more stable and well-delineated fields, as well as a high prevalence of farmland infrastructure, such as irrigation canals, which substantially facilitate field boundary detection. We further exceed the performance of a study in India that used Maxar VHR imagery at sub-meter resolution (Dua et al., 2024). This indicates that 1.5 m SPOT data offer a good tradeoff between landscape complexity, spatial resolution, and accuracy. Our detailed maps of actively used fields provide key information that can inform on crucial socio-economic dynamics (e.g., food production and security, livelihoods, farm consolidation and fragmentation), and environmental outcomes (e.g., deforestation, biodiversity, ecosystem services).

### *Cropland area mapped with VHR data is in line with national agricultural survey*

The map-based active cropland area in *MozFields 2023* adds up to 73,289.88 km². This cropland area corresponds to our unbiased error-adjusted area estimates. For 2023, our reference sample indicates that 76,289 km² ±4,147 km² (9.75% ±0.53%) of the national territory was under active cropland use. The *MozFields 2023* cropland area estimates further agree with the 2023 national agricultural household survey, which documents an overall agricultural area of 69,288 km² (MADER, 2024). The discrepancies (9.1%) can be partly attributed to the omission of large farms (defined as operating on >50 ha of rainfed or >10 ha of irrigated land) in the household survey, which amount to 0.03% of the farms but can be expected to cover a larger proportion of land area (see section National-scale field size distribution reflects dominance of small-scale farming).

### *New cropland maps expose smallholder agriculture in frontier regions*

Overall, the Mozambican agricultural landscape is fragmented and heterogeneous mosaics containing diverse land cover and land use are a key feature in the region (Fig. 1). Half of the actively cropped areas of Mozambique (here defined as 0.05° grid cells with at least 1% of active cropland) have cropland fractions of 11.3% or less. Moreover, 95% of the area has cropland fractions of 51.4% or less.

The spatial distribution of cropland aligns well with the distribution of population (Fig. S2). Sparsely populated regions (< 100 people per 0.05° grid cell, or 3.4 people per km² assuming mean cell area) are characterized by cropland fractions below 10% on average. Cropland fractions increase with population densities (up to 5,000 people per 0.05° grid cell or 170.4 people per km² assuming mean cell area), attesting to agriculture being a key livelihood across the study region. The relationship between population density and cropland fraction reverses in densely populated, mostly urban regions (>5,000 people per 0.05° grid cell) due to constrained land availability.

Compared to publicly available global cropland maps, map-based active cropland area estimates derived from *MozFields 2023* are substantially higher compared to global maps such as ESA World Cover 2021 (Zanaga et al., 2022) (50,943 km²), or GLAD (Potapov et al., 2021) (target year 2019: 40,949 km²). Compared to these products our map identifies an additional 32,341 km² (compared to GLAD 2019) or 22,347 km² (for World Cover 2021) of cropland area.

*MozFields 2023* detects higher cropland fractions in many rural regions and at agriculture-forest frontiers (Fig. 1 subset C, D, and E , Fig. S2). In contrast, we map less active cropland in some of the more populated, accessible and consolidated regions (e.g., Fig. 1 subset B, Fig. S2), which may relate to differences in classification of short-term fallow lands that are typically prevalent in such regions (Rufin et al., 2022). While a direct comparison of these products must be interpreted with caution due to varying class definitions and earlier target years which suppress some of the recent cropland expansion (see Methods section on Validation), our sample-based area estimates confirm the validity of these differences.

*MozFields 2023* reveals *actively cropped regions* - here defined as 0.05° grid cells with >1% active cropland - that were not captured by previous map products (see Methods & Fig. S3 for details). These newly identified agricultural regions are notable, both in terms of area and of the population residing in these landscapes. They cover 77,026 km² (9.5% of the country), often in agricultural frontiers at the margin of consolidated agricultural regions. These regions host ~2.5-3.2 million people (~7-9% of Mozambique's total population) (Tatem, 2017). This population estimate may be considered conservative due to potential underestimation of rural population in global gridded population datasets (Láng-Ritter et al., 2025).

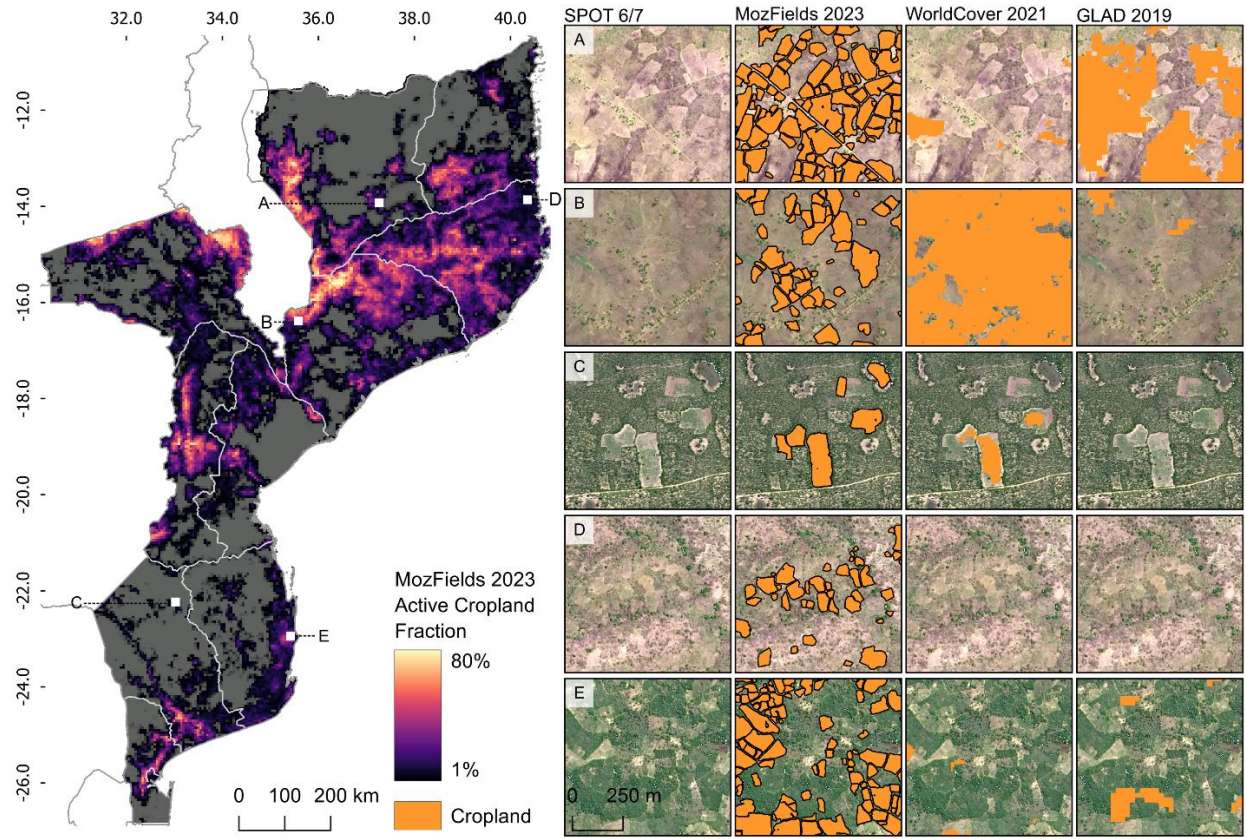

*Fig. 1: Mapped cropland fraction in MozFields 2023 (left). Zoom-ins show Airbus © SPOT6/7 data accessed via Descartes Labs © with MozFields 2023, WorldCover 2021 and GLAD 2019 cropland extent overlaid (right).*

### *National-scale field size distribution reflects dominance of small-scale farming, but large fields cover larger portion of agricultural land than previously estimated*

We mapped 20.9 million (20,933,860) individual fields across Mozambique, which allow robust field size estimation ($R^2$ = 0.86) with a slight tendency for overestimating field size (mean error of 0.037 ha, see Methods and Fig. S4 for more details). Individual field size estimates can be affected by over- or under-segmentation but these appear to smooth out in our aggregated estimates, which we thus consider to be robust. Fields in Mozambique are typically very small, covering on average 0.32 ha, with a standard deviation of 0.56 ha. Half of the fields in Mozambique are smaller than 0.16 ha (median field size), with 83.3% (17,380,213) of the fields being smaller than 0.5 ha and 93.8% (19,642,829) being smaller than 1 ha (Fig. 2).

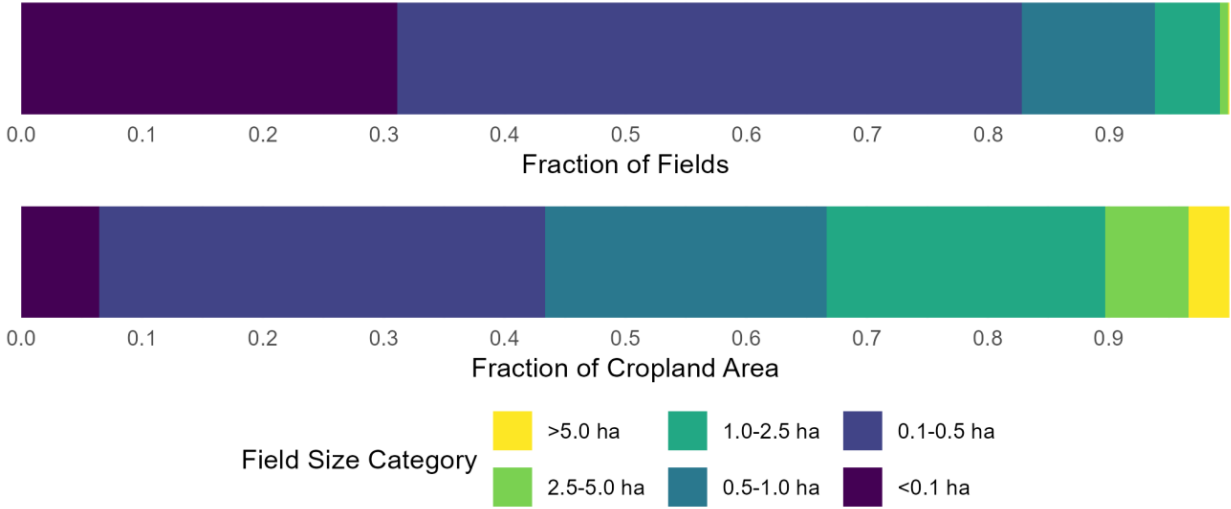

*Fig. 2: Shares of fields and cropland area by field size category at national scale.*

In contrast to the number of fields, the distribution of cropland area per field size category (Fig. 2) shows that fields sized >1 ha cover 33.3%, and fields >2.5 ha cover 10.3% of the Mozambican cropland area. We can relate our mapped field size estimates to the farm size categorization reported in the national agricultural household survey of 2023 (Ministério da Agricultura e Desenvolvimento Rural, 2017). Land-based definitions in the survey categorize small farms as those operating on <10 ha of rainfed land and <5 ha irrigated, medium-size farms operate on 10-50 ha of rainfed, or 5-10 ha of irrigated land, and large farms operate on >50 ha of rainfed, or >10 ha of irrigated land. The survey identified that 98.30% of the farms in Mozambique were small, 1.68% medium and 0.03% large. The average farmland area of the 4.16 million small and medium farms is 1.7 ha, with variation across provinces ranging between 0.6 ha (Maputo) and 2.1 ha (Niassa). From this we can conclude that the larger fields in our inventory, i.e. >1 ha and especially >2.5 ha, are less likely to belong to small and medium farms.

In general, larger field sizes correlate with the presence of larger farms. Moreover, since field size presents a lower bound for farm size, fields >2.5 ha necessarily belong to farms >2.5 ha. The large area proportion of fields >2.5 ha suggests that the importance of medium- and large-scale farms in land use, especially when relying on the commonly admitted thresholds of 2-2.5 ha for smallholder farms, has been underappreciated previously, as also suggested in previous studies based on household surveys across diverse sub-Saharan African contexts (Jayne et al., 2019, 2022).

***The spatial distribution of field size reflects patterns of socio-economic attributes and agricultural development***

Analyses at the 0.05° grid cell level reveal the overall patterns and distribution of field size (Fig. 3). Quantifying mean field size across cropland area reveals that on average, mean field size at the national level is 0.38 ha. Mean field size is lower than 0.5 ha in 83% of the Mozambican territory, and only 1.2% has a mean field size above 1 ha. However, our maps point to local clusters with relatively large (>2.5 ha) mean field size, often surrounding urban agglomerations and agricultural development corridors, but in parts also remote regions.

Field size can be linked to indicators describing the local agricultural context such as cropland fraction and the number of fields, as well as population density and accessibility (Fig. 3). Field size increases with cropland fraction, revealing that consolidated agricultural regions have generally larger fields, but substantial variability exists in regions with mean field size of ~4 ha. The number of fields naturally declines with increasing field size, but the relationship is particularly steep for the smallest field sizes and then saturates. Regions with very small fields are oftentimes linked to floodplain agriculture with high levels of soil moisture, permitting year-round cultivation (Weitkamp et al., 2023). We observe that the smallest field sizes are located in accessible and densely populated regions, for example at the fringes of rural settlements and in suburban contexts. Mean field sizes of 2.5–3.5 ha are found in more remote regions, which is likely related to extensive farming, such as shifting cultivation in forested and sparsely populated landscapes. Contrastingly, the largest mean field sizes of 5 ha are located in more accessible regions, suggesting that these field sizes represent large-scale agricultural operations situated in proximity to transportation infrastructure and markets.

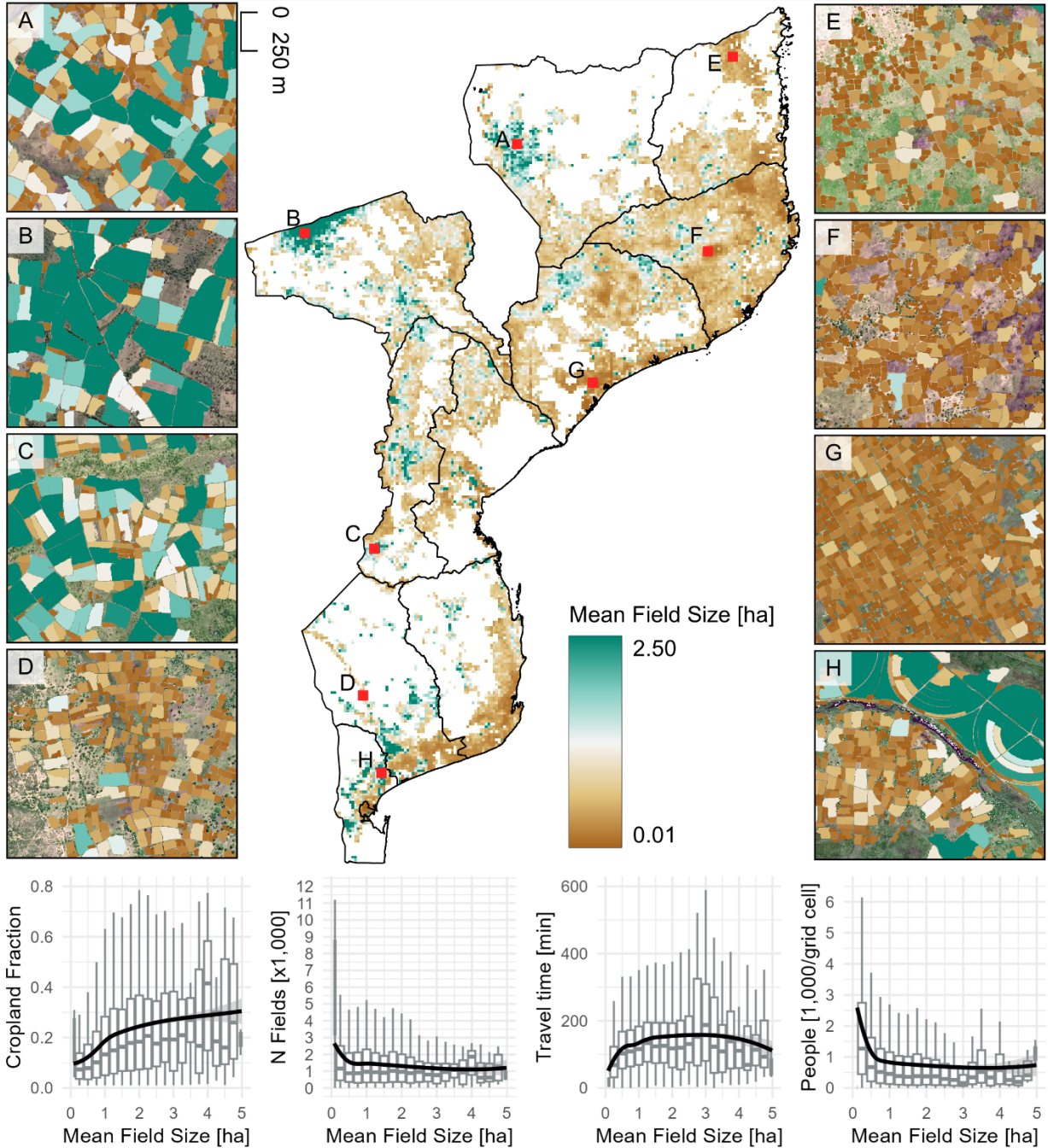

*Fig. 3: Mean field size at 0.05° resolution with zoom-ins to 1.5 m field delineations and Airbus © SPOT6/7 data accessed via Descartes Labs © (top). Statistical distribution of mean field size and cropland fraction, number of fields, travel time to major cities, and population at 0.05° grid cell level (bottom).*

Mapping the fraction of cropland occupied by size category reveals the structure of the agricultural landscape (Fig. 4). Very small fields (<0.1 ha) are more prominent in densely populated and accessible coastal regions. Small fields (0.1-0.5 ha) are the dominant ones, covering around 40% or more of the cropland area in most regions. Fields of intermediate sizes, in the category 0.5-1.0 ha and 1.0-2.5 ha, are prevalent across the country, but particularly in the fringes of agricultural regions. The largest field sizes (2.5-5.0 ha and above 5.0 ha) are prevalent in regions with more productive, commercially oriented and capitalized agriculture. These include the Nacala development corridor from Nampula to Cuamba (Gellert Paris & Rienow, 2023), Gurué district and surroundings (Bey et al., 2020), the Lichinga plateau in Niassa province (Kronenburg García et al., 2022), the Zambian border region in Tete province, or the Chimoio plateau in Manica province (Abeygunawardane et al., 2022).

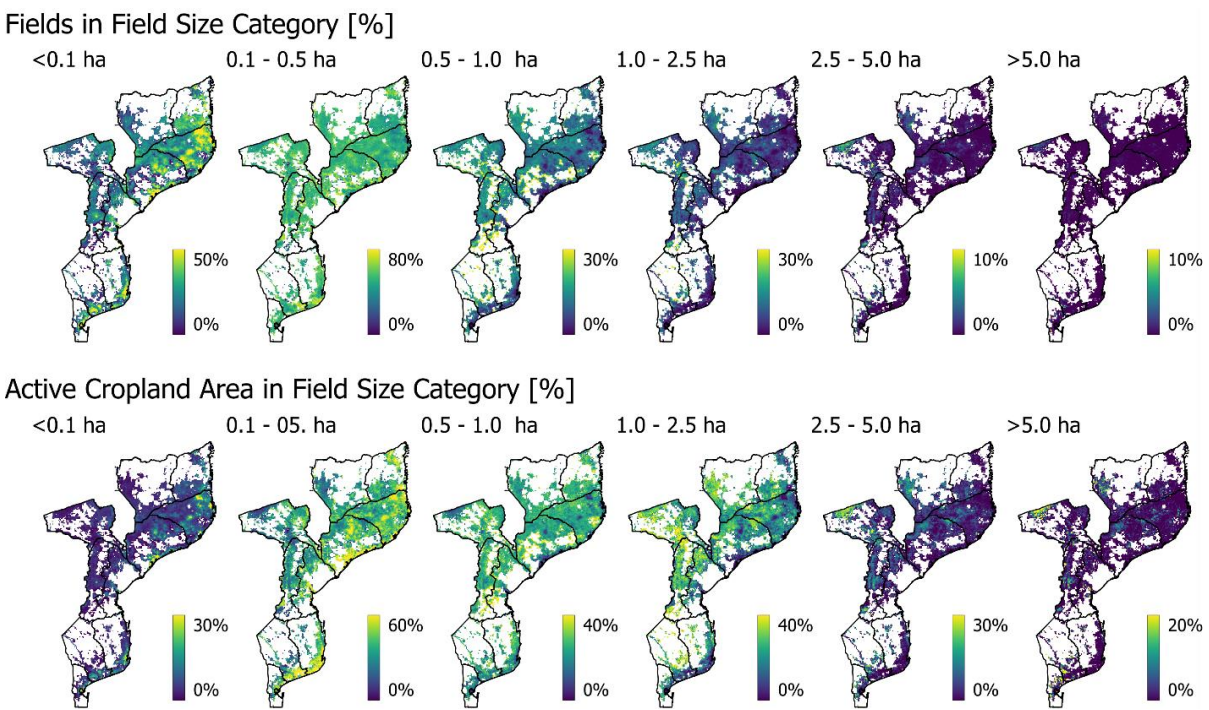

*Fig. 4: Shares of fields by field size category and cropland area by field size category mapped at 0.05° resolution. Note map legends have varying value range.*

### *Higher deforestation is associated with larger field sizes*

We assessed net forest cover change during the period 2010-2020 based on a global land cover and land use change dataset (Potapov et al., 2022). Across Mozambique, 86% of the actively cropped area experienced net forest cover loss (i.e., of >1% of 2010 forest cover), with 50% being subject to substantial (>10%) forest cover loss (Fig. 5). Total net forest cover loss in cropped areas between 2010 and 2020 amounts to 25,964 km². Net forest cover gain (>1%) was observed in 8% of the actively cropped areas, summing up to 383 km². Only 5% of the cropped area did not experience net forest cover change beyond ±1%. Net forest cover loss was notably present in forest-agriculture mosaics with cropland fractions of 0-50% and high forest cover in 2010 (mean forest cover = 49%). High net forest gain was dominantly observed in regions with lower forest cover proportions (mean forest cover = 13%). Some notable outliers of forest cover gain correspond to recently established forestry plantations, such as in the Lichinga plateau in the North and Manica province in central Mozambique (Bey & Meyfroidt, 2021; Chiarella, 2024).

Mean field size varies across net forest change trajectories. Generally, regions with no or low forest change were characterized by the lowest average field size. Larger forest change (both gain and loss) occurred in regions with larger mean field size. High forest loss rates coincide with larger field sizes: regions with 50-100% net forest loss have a mean field size of 1.4 ha as compared to regions with 0-50% net forest loss having a mean field size of 1.0 ha, although the variance in field sizes in low-deforestation regions is very high. Inversely, net loss also varies by field size. Regions with mean field size below 0.5 ha have an average forest cover loss of 13.8%, while field sizes above 2.5 ha have an average forest cover loss of 21.1% in relation to forest cover in 2010.

Most of the deforestation in Mozambique, and sub-Saharan Africa in general, is attributed to smallholders as an indiscriminate category (Branthomme et al., 2023; Masolele et al., 2024). However, smallholder farming involves diverse actors which may produce dominantly for subsistence, engage in local markets, or focus on commercial production. Continuous estimates of field size can contribute to unpacking the nuance in farm types at the sub-national level by linking field size to farm size (Jänicke et al., 2024) and as such may support the identification of medium-scale farms as a separate category of actors which is rapidly emerging across sub-Saharan Africa (Jayne et al., 2019).

Further, regions with higher average field sizes are also those that concentrate much of the commercially-oriented large-scale land investments (Deininger & Xia, 2016; Gellert Paris & Rienow, 2023; Zaehringer et al., 2018). The role of large-scale land investments in deforestation is well acknowledged (Branthomme et al., 2023; Davis et al., 2020), yet the overall footprint of large-scale land investments remains limited in our study region. Our results thus support that disentangling the role of dominantly subsistence-oriented smallholder farms, medium-scale farms, and large-scale investments in these forest cover loss dynamics deserve stronger attention (Wineman et al., 2022).

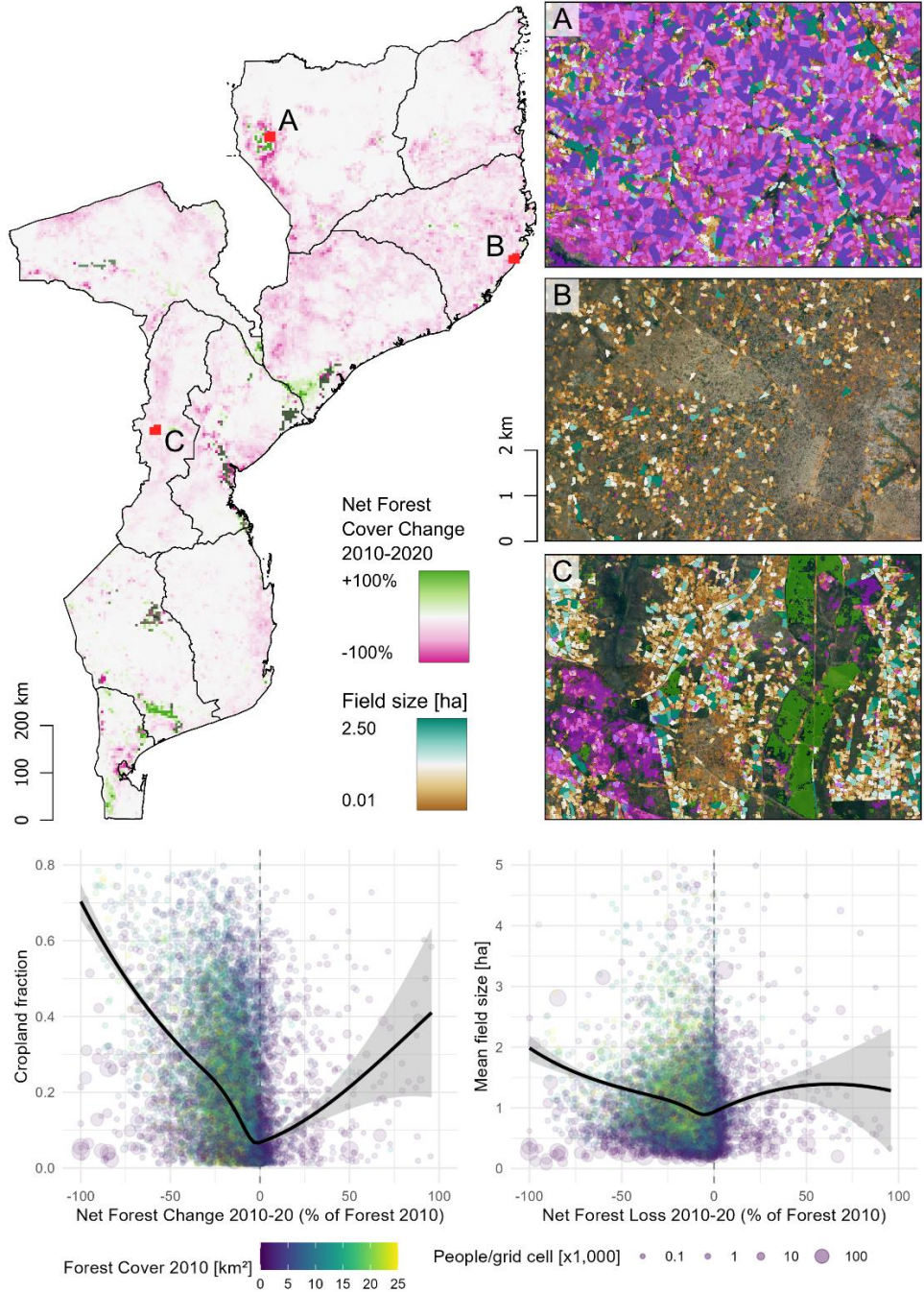

*Fig. 5: Map of net tree cover loss for the period 2010-2020 calculated at 0.05° resolution from Potapov et al. (2022) with selected zoom-ins depicting forest cover change and MozFields 2023 (top). Base imagery is Airbus © SPOT6/7 data accessed via Descartes Labs ©. Scatterplots show the relationship between cropland fraction and net forest cover change (bottom left) and mean field size and forest cover change (bottom right). Point colors relate to baseline forest cover (2010) and point size reflects population density. Trend lines smoothed using locally weighted regressions (loess) weighted by 2010 forest cover.*

# Discussion

### *Adding nuance to the map*

The first set of satellite-based field delineations for Mozambique provides novel insights on the spatial distribution of cropland and field size that support research in the context of SDG 2 (Whitcraft et al., 2019). As opposed to conventional pixel-level mapping efforts, our field-level mapping and associated analyses on field size inform on previously uncaptured agricultural system properties, advancing beyond the mere description of cropland presence. Field size relates to the actors involved in the farming sector, the scale of farming operations, as well as socio-economic and environmental outcomes of agriculture. Even though these relationships are complex and context-dependent, monitoring the spatial distribution of field size can help assessing the status of farming systems across regions (Jänicke et al., 2024). To date, spatially explicit maps of field size are restricted to a global map representing field size categorically (Lesiv et al., 2019). This dataset aligns with our finding that in Mozambique the bulk of the agricultural system harbors very small fields (their definition <0.64 ha). However, the crude field size categories in the global dataset hide the considerable within-category variability. This highlights the added value of our national-level map, which allows to disentangle the variation in mean field size or the proportions of fields or cultivated area falling into a particular size category across the landscape.

### *Persisting image data bottlenecks*

Recent developments in advanced learning techniques increasingly reduce, though do not eliminate, the high quantitative reference data requirements for training deep learning models (Safonova et al., 2023). The potential for geographic transferability and the low reference data requirements when using pseudo-labels for model fine-tuning provide unique opportunities for overcoming current information gaps in smallholder landscapes (Rufin et al., 2024). However, our method leverages large volumes of commercial VHR EO data that are typically not accessible in the context of sustainability-centered research. Our work demonstrates the potential of such endeavors to obtain new information at unprecedented spatial and thematic detail. Nevertheless, even access to the full SPOT6/7 archive did not permit full national coverage at annual intervals. We addressed these challenges by mosaicking images acquired during neighboring years, which helped to mitigate this issue, but small data gaps persist in the study area (<1%). Initiatives to open commercial archives for sustainability-centered research are urgently needed (Nakalembe & Kerner, 2023; Rufin et al., 2025) and future missions providing publicly accessible data at very high spatial resolution should be considered as a way to overcome some of the current knowledge gaps in smallholder farming systems.

### *Uncertainties and remaining challenges*

Our approach allows unprecedented insights into the structure of agricultural landscapes dominated by smallholders, but some uncertainties and challenges remain. These relate to the thematic discrimination between active cropland and the diversity of other land cover types, in particular fallows, the exclusion of tree crops and agroforestry from our analyses, postprocessing of field predictions, as well as the computational requirements linked to post-processing.

Our methods rely on a pre-trained field delineation model, which was not trained for separating cropland from other land cover types (Wang et al., 2022). We introduced this task during fine-tuning of the pre-trained model using a highly constrained training dataset (see

Methods). The predictions of our deep learning model were found to be heavy in commission errors, i.e., false positive predictions. To mitigate this issue, we implemented a machine learning post-processing routine, which on the one hand significantly reduced these errors, but at the expense of increasing omission errors, i.e., false negative predictions. This procedure allowed us to balance error types to avoid bias in cropland extent, as confirmed by the agreement between area estimates obtained from the map and from our independent sample. In parts, the moderate accuracies for our active cropland class stem from the definition excluding short-term fallows, which represented a substantial portion of the errors. For future studies, we recommend further developing the training of non-cropland land cover by mitigating issues related to class imbalance or to rely on ancillary existing or custom-made cropland masks, as typically done in deep learning-based field delineation studies.

We employed machine-learning based post-processing (see Methods on Machine-learning post-processing), which enabled a good balance between error types by filtering falsely identified fields. We note, however, that this procedure was complex in design, requires the collection of additional reference data, and increased omission errors. For this reason, we make the original predictions available to users which can adapt user-defined thresholds to filter fields for a particular area of interest or purpose. For regional to national level analyses on field size indicators, however, we recommend employing the spatially aggregated versions at 0.05° resolution.

In line with most studies on crop field delineation, our definition of fields excludes tree crops and agroforestry systems, which are thus not robustly identified. The importance of cashew, avocado, or macadamia trees in Mozambique, but also cocoa and coffee plantations in other parts of Africa (Masolele et al., 2024), warrants an in-depth study of such systems. In our case, including agricultural systems with high proportions of tree cover has led to systematic confusion with non-agricultural land cover types, which degraded model performance substantially. On a similar note, we were not able to explicitly separate rangelands in this study since the visual resemblance with annual cropland in smallholder farming does not permit a clear-cut distinction. We note, however, that the extent of designated rangeland in Mozambique is very low compared to other parts of southern Africa such as Namibia (Parente et al., 2024).

## Conclusions

Our results demonstrate the potential of deep learning-based field delineation in smallholder contexts based on very-high resolution satellite imagery. Our approach produces unprecedented field inventories and field size maps at very high spatial resolution, allowing to identify very small fields with high thematic accuracy and to create precise reconstructions of field size across large — i.e. national-scale — extents.

Our mapped cropland extent and field size indicators provide novel and crucial insights over (i) the actual extent and spatial distribution of cropland, bringing on the map additional cropland area omitted by existing datasets, in particular in frontier regions, (ii) the very large fraction of cropland area occupied by very small fields, highlighting the challenges faced by most smallholders, (iii) on the other end of the spectrum, the relatively large cropland area occupied by large fields, (iv) the relation between patterns of field size and socio-economic attributes, and (v) the link between large field size and deforestation, emphasizing the often neglected role of medium and large-scale farms in land use dynamics.

Our approach is replicable in other countries, and these data open the way for multiple lines of research and applications to link field size to farm size characterized through household surveys (Jänicke et al. 2024), and to assess the relations between field size and agricultural productivity, socio-economic dynamics and development, and environmental change.

# Materials and methods

*Satellite imagery*

We used SPOT6 and SPOT7 pan-sharpened data at 1.5 m resolution as provided in the Airbus © OneAtlas Living Library, provided on the Descartes Labs © platform. The four-band data includes blue, green, red, and near infrared, and was provided in 8 bits, optimized for visualization purposes. To mitigate low data availability, we created a mosaic comprising observations from the target period 1 January, 2021 through 31 December, 2023 using set of priority rules. We processed the imagery for Mozambique in a tiling scheme. We divided the entire land area of Mozambique into 24,767 tiles with a size of 4096 x 4096 pixels and an overlap of 256 pixels between tiles. The acquisitions included in our data were predominantly acquired in 2023 (70%), followed by 2022 (19%) and 2021 (11%). For some parts of the study region, no imagery was available for the target period, resulting in gaps in the mosaic, accounting for 0.27% of the study area. A detailed description of the image pre-processing steps is in Supplementary Material S1.

*Reference data*

We employed a two-stage sampling design to retrieve our training data (Fig. S5). First, we sampled 4096x4096 pixel tiles using a regular sample with 20 tiles sampling intervals, resulting in 66 tiles for training. For each tile, we collected sparse human annotations for non-cropland and pseudo-labels —i.e., high-confidence predictions of the pre-trained model — for field training. Following Rufin et al. 2024, we used pseudo-labels for geographic domain adaptation. For training, we combined pseudo-labels for fields with human-annotated data for non-cropland. We further collected a set of human field annotations that were used for hyperparameter-tuning during post-processing. Please see Supplementary Material Section S2 for more detailed information on the reference data used for training and hyperparameter tuning.

For sampling our validation locations, we relied on an existing national-level regular sample (Cianciullo et al., 2023) from which we selected a stratified random sample with N=2,500 locations, for which we labeled whether the point is in an actively used field or not at the 1.5x1.5 m pixel level. In 719 samples, a clear decision could not be made due to visual uncertainty, the presence of clouds, or no data in the image mosaic. The final validation sample for the thematic accuracy contained 1,781 labels. From the population of samples in actively used fields, we randomly sampled a subpopulation of 500 locations for which we manually delineated reference field boundaries based on SPOT 6/7 imagery. Owing to visual uncertainty, we could not collect robust reference delineations for 189 out of the 500 locations.

*Model training*

Our field delineation is based on the DECODE framework, encompassing the FracTAL ResUNet for pixel-level semantic segmentation and hierarchical watershed segmentation to derive field instances (see Supplementary Material S3 for a short review of deep learning-based field delineation). We build on the best performing model by Wang et al. (2022), which was trained on sparse labels from France and India for field delineation in Mozambique.

For fine-tuning, we used 3,253 image chips of 256x256 pixel extent which were randomly split into 60% for training, 20% for validation, and 20% for testing. We fine-tuned the model for 100 epochs with loss being calculated only for the labeled pixels, and conducted early stopping based on the maximum Matthews Correlation Coefficient (MCC) of the validation split, which was reached after 32 epochs. The resulting model weights were used for inference across Mozambique, yielding three outputs at the pixel level: the probability of a pixel being within a field, the probability of a pixel being a field boundary, and the within-field distance to the nearest boundary, scaled between 0 and 1. For more details on pre-processing of the inputs and fine-tuning, please see the Supplementary Material S4.

### *Post-processing*

The probability values for cropland extent, field boundary, and within-field distance to the nearest field were converted into field instances using hierarchical watershed segmentation as described in Waldner & Diakogiannis (2020). We determined optimal hyperparameter values using a grid search based on human labels.

After watershed segmentation, we converted raster layers with unique objects into vector polygons. We designed a rule-based approach to merge the overlapping tile-level predictions by accounting for overlapping and incomplete polygons at the tile boundaries. For *MozFields 2023*, we excluded polygons overlapping with waterways contained in the OpenStreetMap database (https://www.openstreetmap.org/).

Fine-tuning the field delineation model and introducing non-cropland as a background class improved the thematic accuracy of our model, but substantial amounts of commission errors, i.e. false detection of fields in non-cropland areas, remained after fine-tuning. While these false positive errors pose challenges for our analyses, they can be corrected in a subsequent post-processing step. As such, commission errors are preferable as compared to omission errors, or false negatives, which are missing fields that cannot easily be recovered in post-processing. For improving the distinction between actual fields and false positives, we designed a workflow based on a Random Forest machine learning model leveraging object-level metrics of field geometry and tile-level attributes. The RF model was used to predict the probability of being an actively used field for each polygon. We found the best balance between errors of commission and omission when thresholding the RF probability at 0.6. For *MozFields 2023,* we provide object-level probability scores that allow users to obtain the original set of predictions from the FracTAL ResUNet and consider their own probability threshold for downstream analyses. Detailed information on the individual post-processing steps is provided in the Supplementary Material S5-S7.

### *Thematic accuracy*

We used the validation sample (N=1,781 consisting of 575 samples for cropland and 1,206 samples for others) and the corresponding stratum weights obtained from the map to produce area-adjusted accuracy estimates as well as error-adjusted class area estimates following the established standards (Olofsson et al., 2014).

Our independent validation reveals robust findings in line with the literature (Table 1). The overall accuracy of the 2023 cropland map is 0.925 ± 0.006, with class-specific user´s and producer´s accuracies above 0.95 for the non-cropland class and above 0.61 for the cropland class. We achieved a very good balance between errors of omission and errors of commission. Error-adjusted class area estimates reveal that the active cropland coverage in Mozambique for 2023 has been 9.75% ± 0.53%.

*Table 1: Area adjusted accuracy and error-adjusted area estimates for the binary active cropland / non-cropland map Table provides class-specific User´s Accuracy (UA), Producer´s Accuracy (PA), as well as Overall Accuracy (OA) with standard errors (SE). Cell values in the confusion matrix correspond to probabilities of encountering the respective combination of map class and reference class in the map.*

|     |          | Reference       |                 |                 |
|-----|----------|-----------------|-----------------|-----------------|
|     |          | Non-crop        | Crop            | UA              |
| Map | Non-crop | 0.8629          | 0.0354          | 0.9606 (0.0055) |
|     | Crop     | 0.0395          | 0.0621          | 0.6111 (0.0271) |
|     | PA       | 0.9562 (0.0043) | 0.6368 (0.0347) |                 |
|     | Area     | 0.9025 (0.0053) | 0.0975 (0.0053) |                 |
|     | OA       | 0.9251 (0.0062) |                 |                 |

We conducted a comparison of our active cropland reference sample for 2023 against WorldCover 2021 and GLAD 2019, which yielded accuracies in a similar range as provided in our assessment for *MozFields 2023*. This again confirms the challenges associated with accurate cropland mapping in complex sub-Saharan African landscapes, as present in Mozambique. Our analyses further confirmed that *MozFields 2023* had the highest accuracies in the active cropland domain (*MozFields 2023* F1: 0.624 ± 0.022, WorldCover 2021 F1: 0.513 ± 0.025, GLAD 2019 F1: 0.372 ± 0.024) with both global products performing worse in terms of omission errors (WorldCover 2021 PA 0.418 ± 0.029; GLAD 2019 PA 0.282 ± 0.023). We want to stress however that a direct comparison with existing map products generated for different target years and varying cropland definitions must be interpreted with utmost caution.

### *Spatial agreement*

We used the manually annotated reference field delineations to assess the spatial agreement of the predictions and reference field delineations as well as field size (Fig. S6). For spatial agreement, we report Intersection over Union (IoU) as mean IoU, median IoU and the fraction of fields with an IoU score above 0.5 (IoU$_{>0.5}$) and 0.8 (IoU$_{>0.8}$). Additionally, we report precision and recall at the object level. We obtained robust results regarding spatial agreement with a mean IoU = 0.756, median IoU = 0.810, IoU$_{>0.5}$ = 0.917, IoU$_{>0.8}$ = 0.533, Dice score = 0.850, Precision = 0.830, Recall = 0.872. For field-level size estimates, we can report RMSE = 0.583, MAE = 0.165, ME = 0.036, as well as R² between observed and predicted field size = 0.865.

### *Spatial aggregation and analyses with ancillary datasets*

For national-scale analyses, *MozFields 2023* and ancillary datasets were aggregated to 0.05° grid cell level. Cell areas of cropped regions varied from 27.54 km² to 30.26 km² across latitudes with a mean of 29.35 km². All analyses on area values at the 0.05° grid cell level account for the varying cell area across latitudes. For example, mapped cropland area per grid cell was determined by multiplying cropland fraction with the total area of the 0.05° grid cell. For our analyses, we defined "actively cropped area" as grid cells with >1% cropland fraction. Regions with net forest change were also considered when change exceeded 1% of the baseline forest cover in both positive and negative directions. For comparisons with population density, we translated people per grid cell to people per km² using the mean cell area of 29.35 km².

We combine indicators derived from *MozFields 2023* with multiple ancillary datasets including data on accessibility defined as travel time to major cities for the year 2015 in minutes by Weiss et al. (2018), population density defined as the number of people residing in each grid cell by Tatem (2017), global land cover information to extract the presence of cropland (Potapov et al., 2021; Zanaga et al., 2022) and net forest cover change (Potapov et al., 2022). For details on the aggregation procedures and product comparisons, please see the Supplementary Material S8.

## Data availability

The *MozFields 2023* field delineations, derivate products, and code will be made accessible through a data sharing repository after peer review. Interested parties may contact the corresponding author for data sharing arrangements prior to publication of this article.

## Acknowledgments

This work was supported by the F.R.S.-FNRS, grant no. T.0154.21 and grant no. 1.B422.24 and the European Research Council (ERC) under the European Union's Horizon 2020 research and innovation program (grant agreement No. 677140 MIDLAND). We gratefully acknowledge the Airbus© SPOT 6/7 data provided by the European Space Agency under the Third-Party Mission Project (PP0091578) and the DescartesLabs© platform. This research contributes to the Global Land Programme (http://glp.earth).

# Supplementary Material

## *S1 Satellite imagery pre-processing*

For each 4096x4096 pixel tile covering Mozambique, we filtered acquisitions with less than 10% cloud cover. If multiple observations were available for the predefined target period (1 January, 2021 – 31 December, 2023), we followed a set of priority rules to select the best observation. We preferred acquisitions with full tile coverage, followed by acquisitions from the target year. If no suitable image was available from the target year, we selected acquisitions with the smallest deviation from the target year. If multiple observations met these criteria, we selected the observation with the lowest cloud cover (although note that we only included acquisitions with cloud cover of 10% or less), and, in case all criteria were met by multiple images, we prioritized acquisitions from later days of the year to maintain a focus on dry season imagery. In case no observation covering the entire tile area were available, we created tile-level image mosaics. For this, images from the most recent acquisition date were mosaicked to achieve complete full tile coverage. In the absence of pixel-level no data and cloud and cloud shadow masks, we used empirically derived thresholds to identify clouds, no data, or cloud shadows. We averaged pixel values across all bands and masked pixels with 0 as no data gaps (0.27%), pixels with an average band average below 10 as shadows (0.76%) and pixels >245 as clouds (1.80%). These masks were used to mask artifacts from the rasterized results.

## *S2 Reference data*

For the human non-cropland annotations, we collected samples in a sparse manner, by roughly delineating 342 larger patches of non-cropland, including forests, savannas, urban agglomerations, rock outcrops, and water surfaces using the annotation tools in QGIS. Following Rufin et al. 2024, we derived pseudo labels from predictions on unlabeled data in the 66 training tiles (Fig. S2). For each tile, we selected the 1% most confident instances with a minimum median confidence of 0.95, yielding a total of 4,127 field polygons. The training polygons for fields and non-cropland were converted into three-band raster labels containing field boundaries, cropland extent, and normalized within-field distance to the nearest boundary as documented in Waldner & Diakogiannis (2020). The 4096x4096 tiles were further subdivided into image chips with a size of 256x256 pixels containing the sparse annotations, which were used as model input for fine-tuning. We further collected human field annotations for hyperparameter-tuning during post-processing. We collected a minimum of five fields in each of the 66 tiles, resulting in 813 fields (average of 12 fields per image).

### *S3 Deep learning-based field delineation*

Field delineation in satellite images is a challenging computer vison task as field boundaries can be simple or complex and field interiors can have spectral characteristics resembling non-field land cover and texture or contrast signatures can be ambiguous. The characteristics of the target agricultural system can influence the field delineation performance. In smallholder-dominated regions, the often complex field geometries are embedded in heterogeneous landscapes, e.g., in savanna environments with mixed vegetation structure and height and the presence of individual trees, which render field delineation challenging.

Deep learning-based field delineation research has gained traction in recent years, with Convolutional Neural Network (CNN) and Transformer models being commonly applied on different satellite data types, most frequently optical satellite data. Most of the architectures are CNN based, particularly U-Net and related architectures are a common choice for field delineation (Hadir et al., 2025).

FracTAL ResUNet is one of these U-Net variants, which leverages the general encoder-decoder architecture with skip-connections (Ronneberger et al., 2015). The skip-connections enable maintaining detailed semantic segmentations and detailed feature representation throughout the encoding and decoding process. Additionally, the FracTAL ResUNet uses residual blocks typical to ResNet architectures and complements them with self-attention layers called FracTAL units (Waldner et al., 2021; Waldner & Diakogiannis, 2020). The self-attention mechanism represents a key element commonly used in more recent Transformer architectures, making the FracTAL ResUNet a competitive model for field delineation tasks. Recent comparison studies repeatedly highlight the FracTAL ResUNet to perform on par or better than other field delineation algorithms in Germany (Tetteh et al., 2023), or France (Hadir et al., 2025 mean IoU of 0,81), where it outperformed a low-rank-based fine-tuned Segment-Anything (Lora-SAM), which is a complex Transformer based state-of-art model. The competitive performance of the FracTAL ResUNet as well as openly available pretrained model weights that can be leveraged using domain adaptation and transfer learning make it very suitable for our research (Rufin et al., 2024; Wang et al., 2022).

### *S4 Model training*

The 256x256 pixel image chips and corresponding labels (N=16,898) were imbalanced with respect to general label coverage and the fraction of chips with field boundary labels, with an overrepresentation of chips containing no fields. To reduce imbalance in the training dataset, we removed 95% of the image chips containing non-field but no field boundary pixels. The final training dataset contained 3,253 image chips with sparse labels which were randomly split into 60% for training, 20% for validation, and 20% for testing. Compared to conventional data splits we used more data for validation and testing to obtain a stable baseline for evaluating our fine-tuning procedure.

### *S5 Watershed segmentation*

The probability values for cropland extent, field boundary, and within-field distance to the nearest field were converted into field instances using hierarchical watershed segmentation as described in Waldner & Diakogiannis (2020). The optimal hyperparameter values were determined using a grid search based on human labels, optimizing for mean IoU (mIoU). This also allowed a comparison against the performance of the pre-trained model, demonstrating a substantial performance increase of +0.14 mIoU after fine-tuning. The highest mIoU and $IoU_{>0.5}$ scores served to identify the optimal thresholds for cropland extent (t_ext = 0.2) and field boundary probability (t_bnd = 0.2), which we used for the watershed segmentation (Fig. S7).

### *S6 Combining overlapping predictions*

After watershed segmentation, we converted raster layers with unique objects into vector polygons. We designed a rule-based approach to merge the overlapping tile-level predictions by accounting for overlapping and incomplete polygons at the tile boundaries. In the case of overlapping complete polygons, we selected the polygon with the highest prediction confidence if one of the two conditions applied: (1) the centroid of the smaller polygon fell into the larger polygon, or (2) the overlap area of both polygons was larger than 30%. This ensured that polygons that only overlapped at the edges, e.g. due to shifts in the viewing angle, were not falsely removed. In the case of incomplete polygons at the tile boundaries, we removed incomplete polygons that intersected with the outer tile border assuming that the complete polygon was included in the prediction of the adjacent tile. We dissolved polygons that intersected both the inner and outer border of the overlap area if the centroid of the smaller polygon laid within the larger polygon, or the overlap area was larger than 30%. In the case of remaining polygons that overlapped only slightly (overlap <30%), we differenced the geometries of intersecting polygons. The geometry of the polygon with the lower prediction confidence was then modified to remove the overlap with the higher confidence prediction. We thereby prioritized fields based on prediction confidence so that fields with the highest prediction confidence were not adjusted.

We ran the routine for 3 by 3 tiles as a moving window across the study region, thereby combining the predictions into a larger 3x3 tile. We repeated the merging routine with the resulting 3x3 tiles to clean the predictions at the edges of the 3x3 tiles. After cleaning and dissolving overlapping polygons, we added an attribute indicating whether the polygons intersect a waterway or not. For *MozFields 2023*, we excluded polygons overlapping with waterways. We used only the linear features of the OpenStreetMap water product because the extent of lakes and rivers differs strongly between wet and dry season and some fields are precisely cultivated in riverbeds. As a result, we obtained ~2,700 files containing vectorized geometries. Each file covers ~17.664 km vertical and horizontal extent without overlaps or incomplete predictions.

### *S7 Machine-learning post-processing*

For training a machine learning model to identify commission errors, we sampled a set of 2,500 reference fields using a two-stage sampling design. We first randomly sampled 500 tiles across the study region and subsequently conducted a random sample of n=5 polygons within each tile. We labeled each polygon as field or non-cropland based on SPOT 6/7 imagery. Polygons dominantly covering active cropland were labeled as field, while polygons dominantly covering non-cropland were labeled as non-cropland. The labeling was conducted irrespective of the segmentation quality, meaning that even inaccurate delineations were included as field, as long as their area covers mostly active cropland.

We split these labeled polygons into training (n=1,511) and testing (n=1,000) and trained a Random Forest (RF) machine learning classification algorithm with 250 trees and trying 3 variables at each split. We included predictor variables representing geometric properties of the polygons, and multi-scale indicators on cropland presence, polygon size, and classification confidence. Specifically, we used object-level fractal dimension and circumference-area-ratio as common parameters used to describe the shape of agricultural parcels (Oksanen, 2013). For cropland presence, we calculated tile-level cropland fraction and 3x3 tile-level cropland fraction from the GLAD cropland product for 2019 (Potapov et al., 2021). For polygon size we calculated mean polygon size at the 3x3 tile-level, and the individual polygon size relative to the mean field area at the 3x3 tile-level. For classification confidence, we used object-level median prediction probability, and object-level median prediction probability relative to the mean at the 3x3 tile-level.

Consecutively, the RF model was used to predict the probability of being an actively used field for each polygon. The approach achieved a substantial reduction of commission errors with moderate increases of omission errors. We found the best balance between errors of commission and omission when thresholding the random forest probability at 0.6.

*S8 Ancillary datasets and analyses*

For national-scale analyses, *MozFields 2023* was aggregated to 0.05° grid cell level. We rasterized field polygons at the native resolution (0.0000135°, or 1.5 m) to calculate metrics on cropland fraction, field size, number of fields, number of fields per size category, and fraction of cropland area belonging to each field size category. For cropland fraction, field polygons were rasterized to a binary representation of fields and non-fields. We used mean aggregation on this binary dataset to calculate the fraction of cropland per grid cell. Data gaps and cloud or shadow cover in the image mosaic were considered during this process and in case present, values indicate the fractions relative to the mapped area. For field size, we rasterized the size information of each field and aggregated this information to the 0.05° grid cell level with a mean aggregation excluding non-cropland areas. For the number of fields and the fraction of fields per size category, we rasterized field centroids with value 1 and performed sum operations at the 0.05° grid cell level. For size class-specific fractions we rasterized only the centroids of the size class of interest and divided by the total number of fields per grid cell.

We combine indicators derived from *MozFields 2023* with ancillary datasets including data on accessibility defined as travel time to major cities for the year 2015 in minutes by Weiss et al. (2018), population density defined as the number of people residing in each grid cell by Tatem (2017), global land cover information to extract the presence of cropland (Potapov et al., 2021; Zanaga et al., 2022) and net forest cover change (Potapov et al., 2022). Travel time was aggregated to 0.05° using a mean resampling function, while population (people per grid cell) was aggregated using a sum resampling to sum all pixels falling into each 0.05° grid cell. In the case of WorldCover 2021, we created binary representations of cropland presence (with 0 as non-cropland and 1 as cropland) which were aggregated pixel-wise using a mean aggregation function. Pixel values thus represent the cropland fractions per grid cell. To create a net forest cover change dataset, we first converted all classes with tree cover >5 m to a binary forest representation for 2010 and 2020 and calculated the difference between both time periods as net change. We measured net change in % relative to the 2010 forest cover.

When comparing actively cropped regions in *MozFields 2023* compared to GLAD 2019 and World Cover 2021, we ensured that the newly mapped regions are not a consequence of higher rates of commission errors by checking the spatial error distribution in our reference sample across categories (Fig. S3). This analysis confirmed that most of the mapping errors (95%) occur in regions outside of the newly added croplands, suggesting errors related to short-term fallow in regions with higher cropland fractions. The proportion of correctly mapped points is higher in the newly added cropland (95.4%) compared to the remaining regions (73.3%). Furthermore, the distribution of commission versus omission errors remains balanced in the newly added croplands with 48% omission and 52% commission errors (Fig. S4).

*Supplementary Figures*

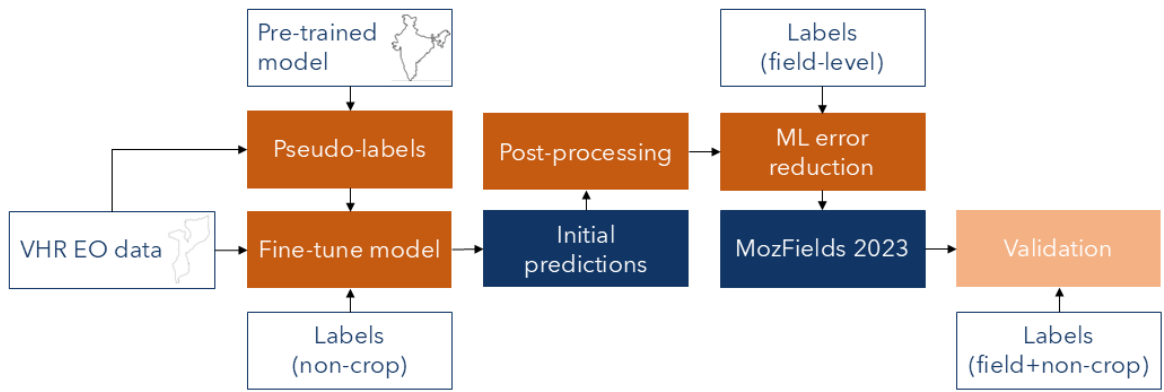

*Fig. S1: Workflow for national-scale field delineation using transfer learning and pseudo-labels.*

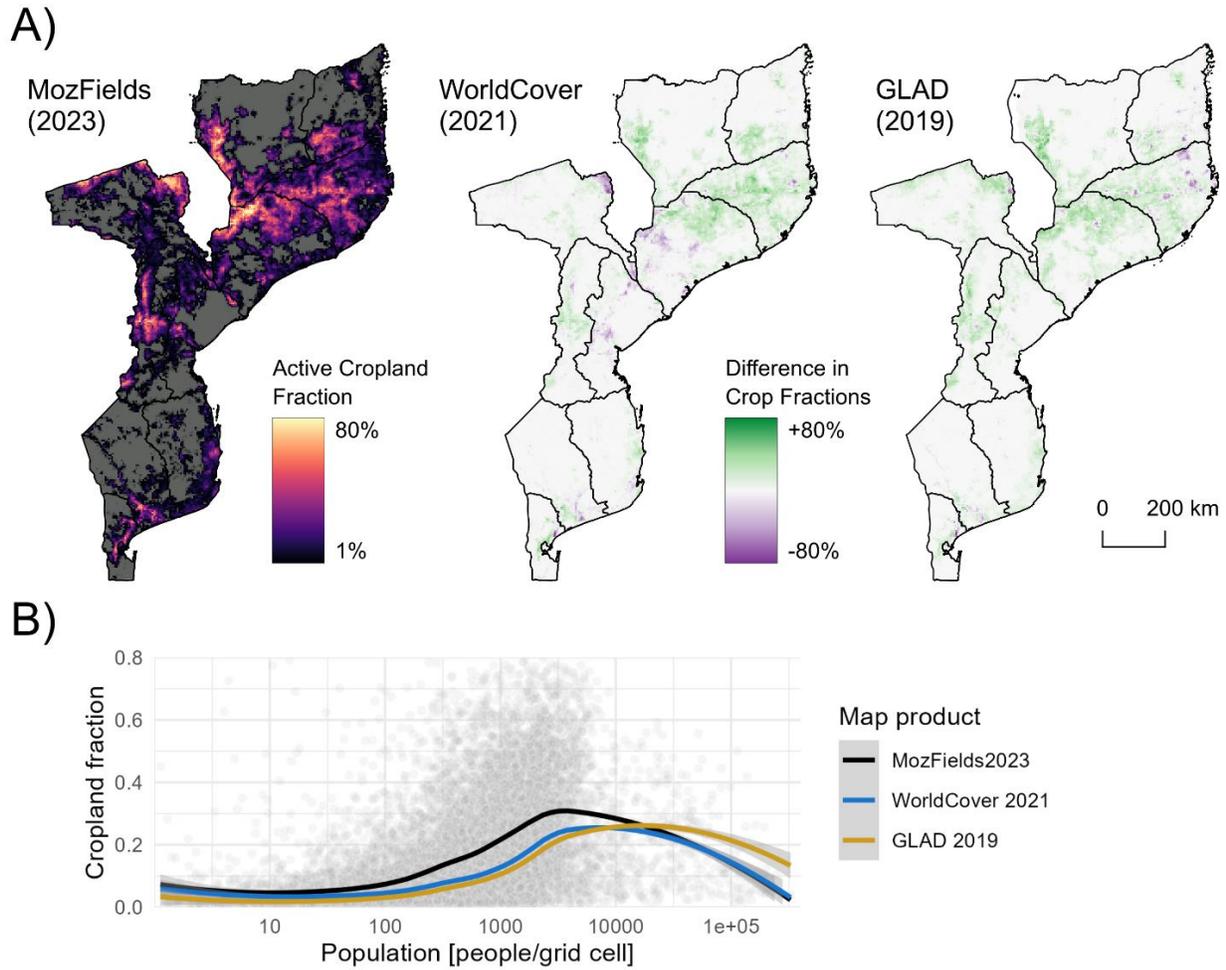

*Fig. S2: Difference in crop fractions between Mozfields 2023 and WorldCover 2021 and Mozfields 2023 and GLAD 2019 (A) Green color indicates higher cropland fraction in MozFields 2023 compared to the other datasets. Relationship between cropland fraction and population for MozFields (black line and grey points), WorldCover (blue line) and GLAD (orange line), smoothed using locally weighted regressions (loess) (B).*

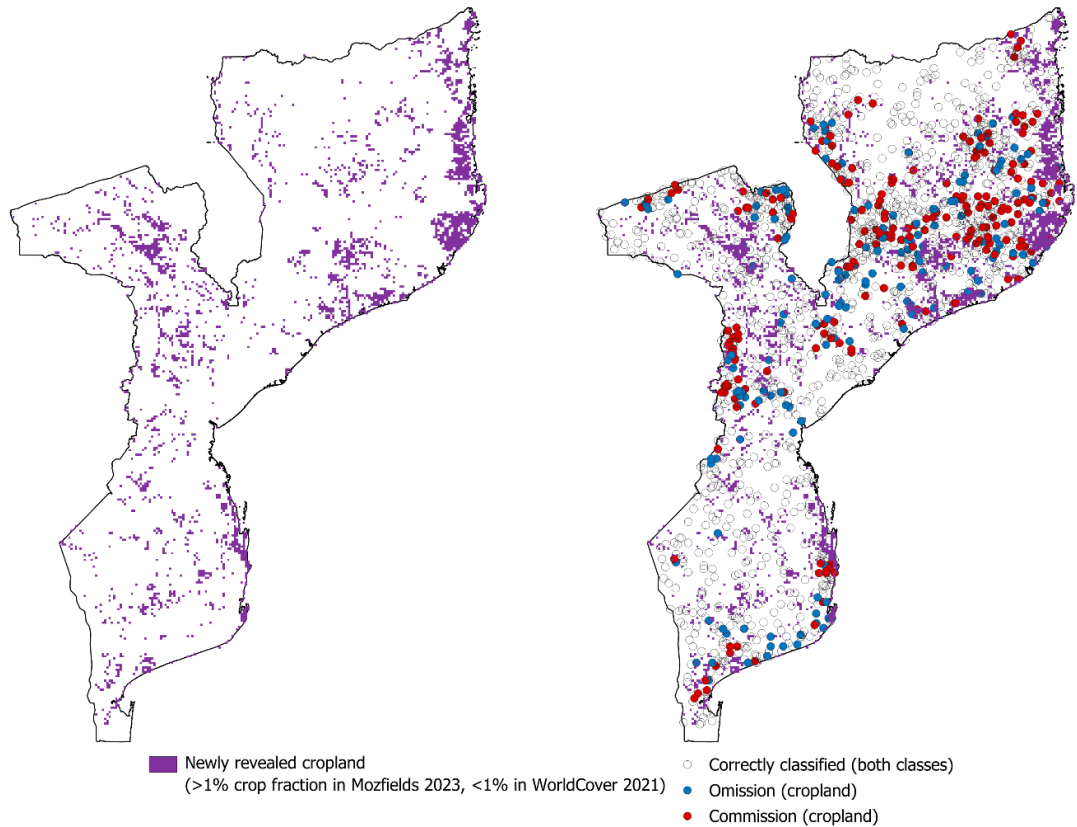

*Fig. S3: Map of newly mapped cropland regions added by MozFields 2023 in comparison to WorldCover 2021 (left) and overlaid reference sample for validation with correctly classified samples (hollow circles), errors of omission (blue) and commission (red).*

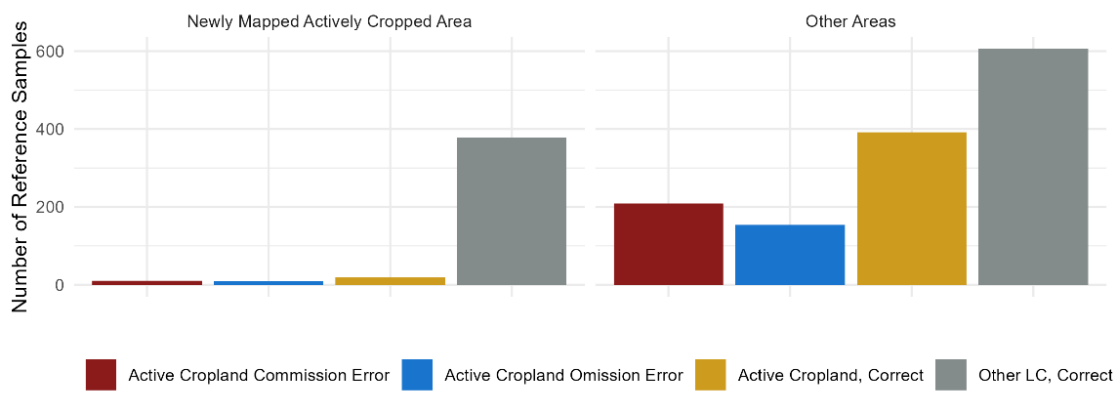

*Fig. S4: Bar chart of the number of reference samples for active cropland and other land cover that are correctly or erroneously classified for the newly mapped cropland areas and the remaining areas.*

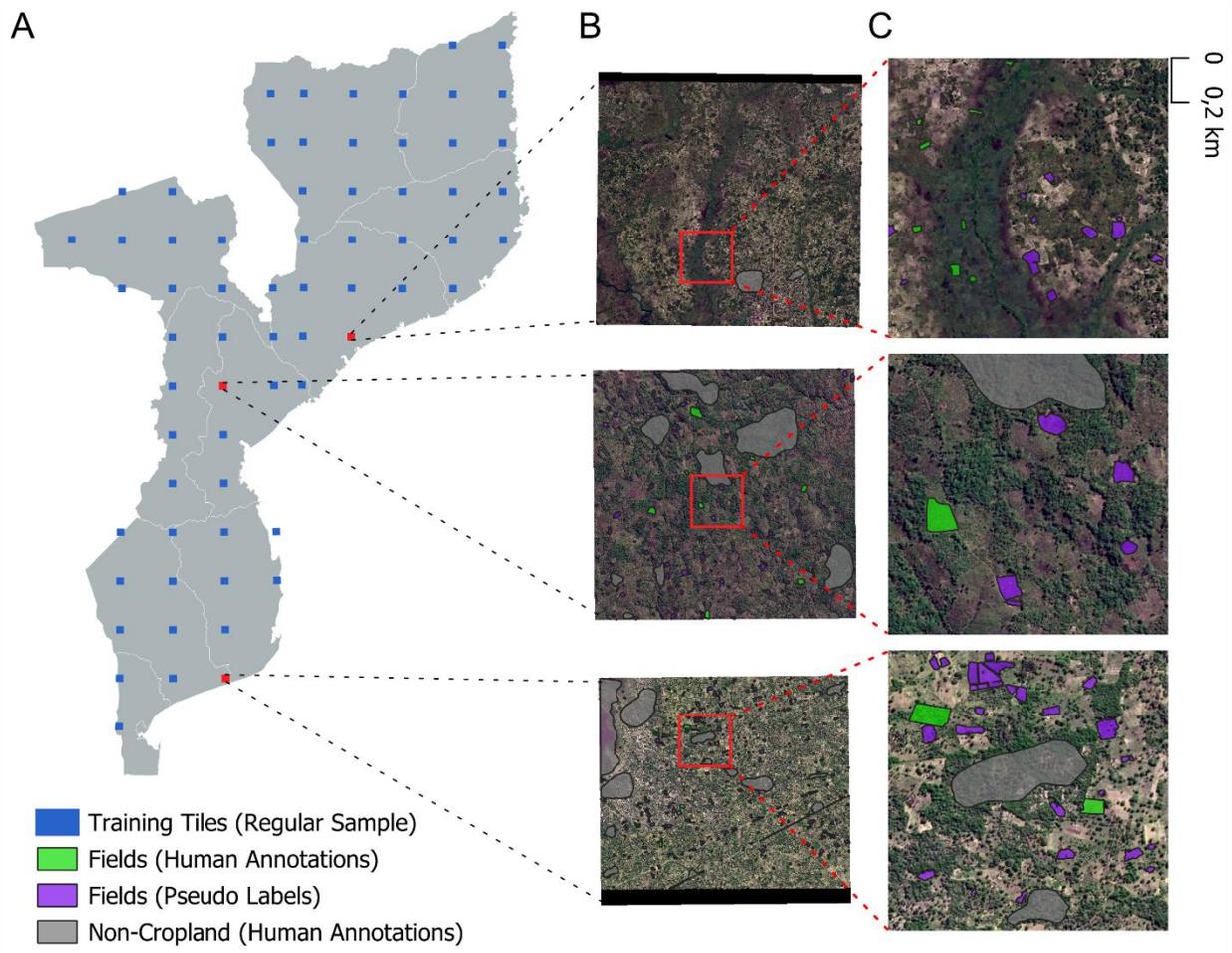

*Fig. S5: Two-stage sampling scheme for training data with tile-level regular sampling yielding 66 training tiles (A). Examples of tile-level reference data (B) with zoom ins (C) showing Airbus © SPOT6/7 data accessed via Descartes Labs ©. Human annotations for fields (green) were used for hyperparameter tuning. Pseudo-labels for fields (purple) and human annotations for non-cropland (grey) were used for model fine-tuning.*

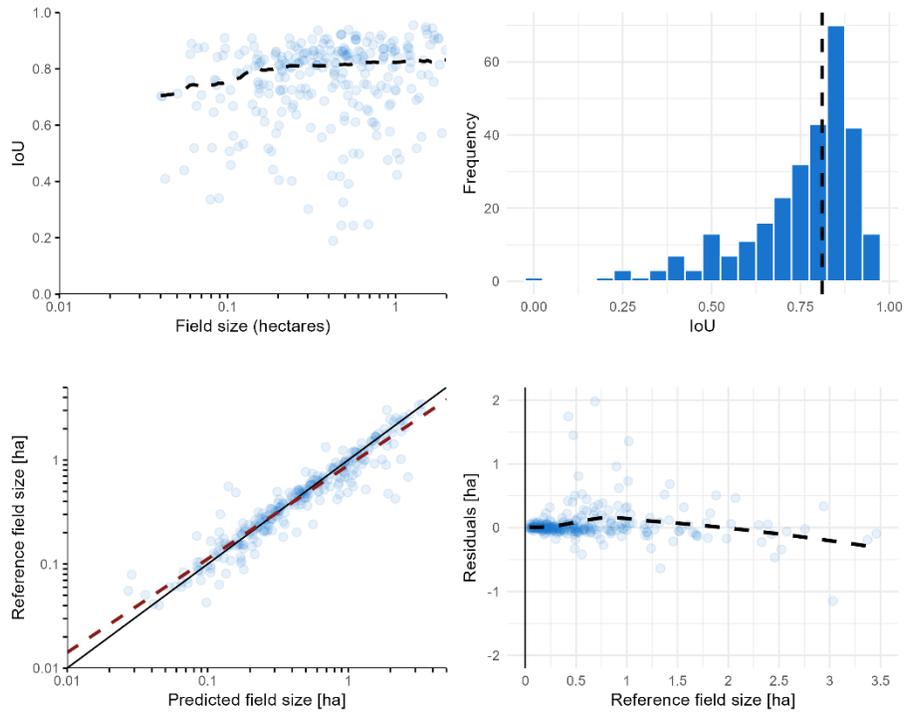

*Fig. S6: Field-level assessment of IoU and field size estimates. IoU scores against field size with black dashed line representing rolling median IoU (top left) and histogram of IoU scores across population of reference fields with black dashed line representing median IoU (top right). Scatterplot of observed versus predicted field size with linear relationship as dashed red line and 1:1 fit as black line (bottom left). Residuals between observed and predicted field size against observed field size with black dashed line representing locally-weighted smoothing (bottom right).*

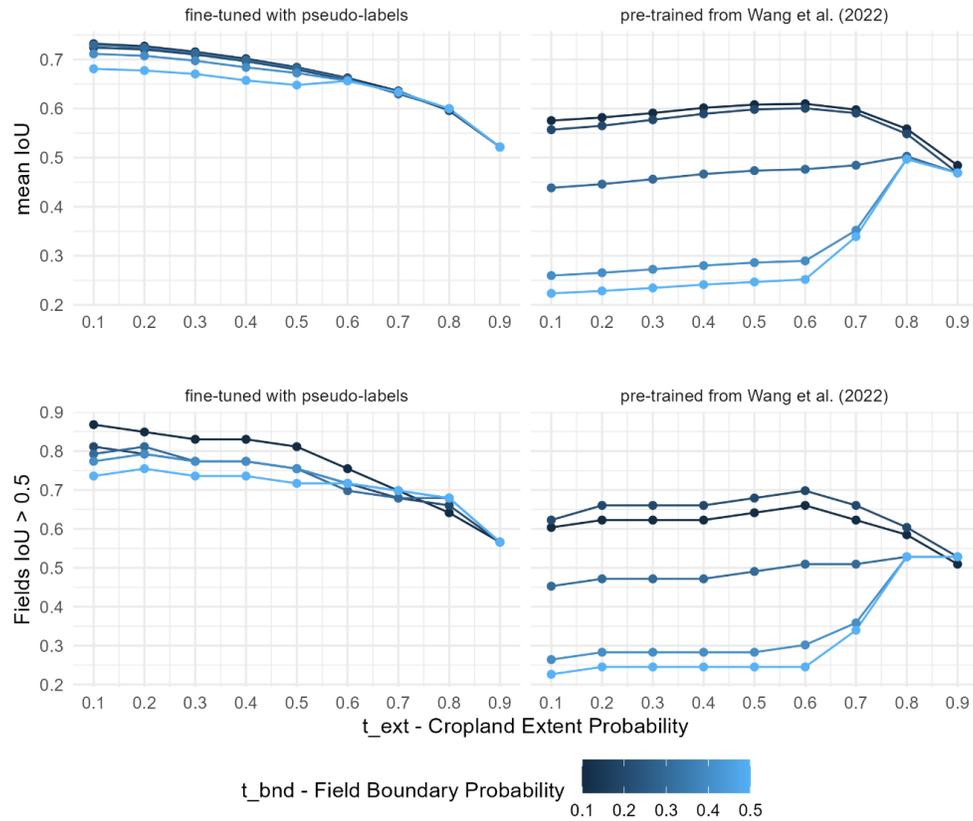

*Fig. S7: Sensitivity analyses of hierarchical watershed segmentation. Mean IoU and IoU 50 scores were calculated across cropland extent and boundary thresholds to identify optimal hyperparameters. Models fine-tuned with pseudo-labels (left column) outperform pre-trained models without fine-tuning (right column).*

Mozambique (2001–2016). *Rendiconti Lincei. Scienze Fisiche e Naturali*. https://doi.org/10.1007/s12210-023-01133-9

Davis, K. F., Koo, H. I., Dell'Angelo, J., D'Odorico, P., Estes, L., Kehoe, L. J., Kharratzadeh, M., Kuemmerle, T., Machava, D., Pais, A. de J. R., Ribeiro, N., Rulli, M. C., & Tatlhego, M. (2020). Tropical forest loss enhanced by large-scale land acquisitions. *Nature Geoscience*, *13*(7), 482–488. https://doi.org/10.1038/s41561-020-0592-3

Deininger, K., & Xia, F. (2016). Quantifying Spillover Effects from Large Land-based Investment: The Case of Mozambique. *World Development*, *87*(2), 227–241. https://doi.org/10.1016/j.worlddev.2016.06.016

Dua, R., Saxena, N., Agarwal, A., Wilson, A., Singh, G., Tran, H., Deshpande, I., Kaur, A., Aggarwal, G., Nath, C., Basu, A., Batchu, V., Holla, S., Kurle, B., Missura, O., Aggarwal, R., Garg, S., Shah, N., Singh, A., … Talekar, A. (2024). *Agricultural Landscape Understanding At Country-Scale* (arXiv:2411.05359). arXiv. https://doi.org/10.48550/arXiv.2411.05359

Estes, L. D., Ye, S., Song, L., Luo, B., Eastman, J. R., Meng, Z., Zhang, Q., McRitchie, D., Debats, S. R., Muhando, J., Amukoa, A. H., Kaloo, B. W., Makuru, J., Mbatia, B. K., Muasa, I. M., Mucha, J., Mugami, A. M., Mugami, J. M., Muinde, F. W., … Caylor, K. K. (2022). High Resolution, Annual Maps of Field Boundaries for Smallholder-Dominated Croplands at National Scales. *Frontiers in Artificial Intelligence*, *4*, 744863. https://doi.org/10.3389/frai.2021.744863

Gellert Paris, R., & Rienow, A. (2023). Using geospatial data to identify land grabbing. Detecting spatial reconfigurations during the implementation of the Nacala Development Corridor

*Annual grassland class and extent maps at 30-m spatial resolution (2000 – 2022)* (Version v1) [Dataset]. Zenodo. https://doi.org/10.5281/ZENODO.13890400

Pendrill, F., Gardner, T. A., Meyfroidt, P., Persson, U. M., Adams, J., Azevedo, T., Bastos Lima, M. G., Baumann, M., Curtis, P. G., Sy, V., Garrett, R., Godar, J., Goldman, E. D., Hansen, M. C., Heilmayr, R., Herold, M., Kuemmerle, T., Lathuillière, M. J., Ribeiro, V., … West, C. (2022). Disentangling the numbers behind agriculture-driven tropical deforestation. *Science*, *377*(6611), 9267. https://doi.org/10.1126/science.abm9267

Potapov, P., Hansen, M. C., Pickens, A., Hernandez-Serna, A., Tyukavina, A., Turubanova, S., Zalles, V., Li, X., Khan, A., Stolle, F., Harris, N., Song, X.-P., Baggett, A., Kommareddy, I., & Kommareddy, A. (2022). The Global 2000-2020 Land Cover and Land Use Change Dataset Derived From the Landsat Archive: First Results. *Frontiers in Remote Sensing*, *3*, 856903. https://doi.org/10.3389/frsen.2022.856903

Potapov, P., Turubanova, S., Hansen, M. C., Tyukavina, A., Zalles, V., Khan, A., Song, X.-P., Pickens, A., Shen, Q., & Cortez, J. (2021). Global maps of cropland extent and change show accelerated cropland expansion in the twenty-first century. *Nature Food*, *327*, 812. https://doi.org/10.1038/s43016-021-00429-z

Ricciardi, V., Mehrabi, Z., Wittman, H., James, D., & Ramankutty, N. (2021). Higher yields and more biodiversity on smaller farms. *Nature Sustainability*, *4*(7), 651–657. https://doi.org/10.1038/s41893-021-00699-2
52

Safonova, A., Ghazaryan, G., Stiller, S., Main-Knorn, M., Nendel, C., & Ryo, M. (2023). Ten deep learning techniques to address small data problems with remote sensing. *International Journal of Applied Earth Observation and Geoinformation*, *125*, 103569. https://doi.org/10.1016/j.jag.2023.103569

Samberg, L. H., Gerber, J. S., Ramankutty, N., Herrero, M., & West, P. C. (2016). Subnational distribution of average farm size and smallholder contributions to global food production. *Environmental Research Letters*, *11*(12), 124010. https://doi.org/10.1088/1748-9326/11/12/124010

Sitko, N., & Chamberlin, J. (2015). The Anatomy of Medium-Scale Farm Growth in Zambia: What Are the Implications for the Future of Smallholder Agriculture? *Land*, *4*(3), 869–887. https://doi.org/10.3390/land4030869

Tatem, A. J. (2017). WorldPop, open data for spatial demography. *Scientific Data*, *4*(1), 170004. https://doi.org/10.1038/sdata.2017.4

Tetteh, G. O., Schwieder, M., Erasmi, S., Conrad, C., & Gocht, A. (2023). Comparison of an Optimised Multiresolution Segmentation Approach with Deep Neural Networks for Delineating Agricultural Fields from Sentinel-2 Images. *PFG – Journal of Photogrammetry, Remote Sensing and Geoinformation Science*, *91*(4), 295–312. https://doi.org/10.1007/s41064-023-00247-x

Tong, X., Brandt, M., Hiernaux, P., Herrmann, S., Rasmussen, L. V., Rasmussen, K., Tian, F., Tagesson, T., Zhang, W., & Fensholt, R. (2020). The forgotten land use class: Mapping of